# A Hybrid Feature Selection and Construction Method for Detection of Wind Turbine Generator Heating Faults

Ayse Gokcen Kavaz
Huawei Turkey R&D Center
ayse.gokcen.taskiner@huawei.com

Burak Barutcu
Istanbul Technical University, Energy Institute
barutcub@itu.edu.tr

**Abstract**

Preprocessing of information is an essential step for the effective design of machine learning applications. Feature construction and selection are powerful techniques used for this aim. In this paper, a feature selection and construction approach is presented for the detection of wind turbine generator heating faults. Data were collected from Supervisory Control and Data Acquisition (SCADA) system of a wind turbine. The original features directly collected from the data collection system consist of wind characteristics, operational data, temperature measurements and status information. In addition to these original features, new features were created in the feature construction step to obtain information that can be more powerful indications of the faults. After the construction of new features, a hybrid feature selection technique was implemented to find out the most relevant features in the overall set to increase the classification accuracy and decrease the computational burden. Feature selection step consists of filter and wrapper-based parts. Filter based feature selection was applied to exclude the features which are non-discriminative and wrapper-based method was used to determine the final features considering the redundancies and mutual relations amongst them. Artificial Neural Networks were used both in the detection phase and as the induction algorithm of the wrapper-based feature selection part. The results show that, the proposed approach contributes to the fault detection system to be more reliable especially in terms of reducing the number of false fault alarms.

## 1. Introduction

Condition monitoring and fault detection are amongst the significant topics in wind turbine research. As the size and number of wind turbines continue to grow in line with the global renewable energy targets, fault detection of wind turbines became even more important. Faults generally lead to downtimes in wind turbine operation and result in a decrease in the amount of energy conversion. Moreover, unpredicted faults can have detrimental effects on other parts of wind turbines which contribute to decrease in lifetime of overall system. Therefore, early detection and isolation of wind turbine faults are required.

Fault detection approaches can be investigated in two main classes which are model-based and data driven methods. In model-based fault detection, an explicit mathematical model of system is generated and outputs of the real system are evaluated comparing to responses of the mathematical model. Model-based methods have the advantage of not requiring high frequency data, however their performance highly depends on the accuracy of the mathematical model which is difficult to build in real world applications and the model-based applications have a limited capability in supplying the details about wind turbine faults [1]. Data-driven methods are based on the analysis of measurements collected from the corresponding system. In the early data driven fault detection studies, multivariate statistics were widely used. With the advancements of intelligent algorithms, machine learning methods became more widespread. Collection of wind turbine data for fault detection can be done in various ways. First of them is to collect data from specifically mounted sensors [2]–[6]. As these sensors are mounted for condition monitoring and fault detection aims, useful high frequency data can be obtained. However, this approach brings additional costs. The other alternative is to use data from Supervisory Control and Data Acquisition (SCADA) system [7]–[12]. The main advantage of this alternative is that SCADA is a built-in part of most modern wind turbines. Therefore, additional hardware costs are not required. However, SCADA systems were not initially built for fault detection aims, so there are imperfections in data such as high proportion of missing values. Moreover, the data output interval of wind turbine SCADA systems is generally 10 min. This low output rate results in the loss of high frequency data which is very useful in fault detection studies. To reduce the disadvantages of using SCADA data, intelligent data processing approaches are required. In this paper, a data-driven method using SCADA data was realized by implementing feature selection techniques along with Artificial Neural Networks (ANN).

SCADA data set consists of many measurements such as wind turbine operation, wind speed characteristics, temperature values and fault information. In addition to these directly collected data, a feature construction step was conducted to have more indications on incipient faults. By this way, the number of the input features increased even further. Feature selection is to identify a subset of relevant features from the overall feature set which is a compulsory process in machine learning applications involving moderate and high number of input features [13]–[19]. It brings many advantages such as preventing overfitting that can be caused by large number of features, reducing computational burden and training time, increasing accuracy of model. Feature selection process can be employed by different approaches namely filter, wrapper and embedded techniques. Filter approaches evaluate features without utilizing any classification algorithm. They rank features independently based on a selected criteria [13], [20]. Wrapper methods select and evaluate a subset of features together and search for the best subset describing the model [21]. In embedded approaches, the selection is a part of the learning process [22]. In this paper, a hybrid feature selection method was employed. In the first step, various filter methods were applied to find out and exclude the features that are non-discriminant. The remaining features which are informative on faults were evaluated in a wrapper-based approach to get the knowledge about mutual relations or additional redundancies. By combining these two approaches, it was aimed to benefit from the advantages of both. Filter methods are practical in large data sets in terms of training time and complexity. However, they are not able to evaluate mutual dependencies between features. Therefore, after using the filter approach as a pre-processing step, wrapper method was employed to eliminate redundancies and find subsets based on evaluating mutual relations.

The type of wind turbine faults observed in this paper are generator heating faults. They are non-fatal but frequently occurring wind turbine faults. These type of faults show less indications than fatal faults but are one of the important reasons of long downtime durations. The success of detecting frequent/non-fatal faults are required to be increased especially in terms on reducing false alarms [23]–[25].

The layout of this paper is as follows; in Section 2, the data were described in detail. In Section 3, the hybrid feature selection method was presented. In Section 4 results were presented and in Section 5 conclusion was given, respectively.

## 2. Data Description

The data used in this study were collected from a three-bladed horizontal axis wind turbine with a rated power of 900 kW. They were collected from 01.01.2105 to 31.12.2015. Original and generated features are described in this section.

### 2.1 Collected data

Original features are the measurements and information directly collected by the data collection system. Wind, temperature, operational and status data are the original features of this system. First three types of them have a 10 min sampling period, whereas a new status data is produced when the status of the turbine changes. This situation results in a difference in the number of samples. To match the status data with other types of information, a status label was generated for each 10 min time instance.

**Wind Data:** Wind speed measurements consist of minimum, maximum and average values for each 10-min interval.

**Temperature Data:** There are 9 temperature sensors mounted on different parts of the turbine. Locations of temperature sensors are presented in Table 1.

Table 1. Temperature sensors.

| Location of Temperature Sensors |
|---|
| Generator stator |
| Generator rotor |
| Nacelle |
| Front hub bearing |
| Rear hub bearing |
| Nacelle control cabinet |
| Control cabinet |
| Tower |
| Transformer |

**Operational Data:** SCADA systems supply various operational data to monitor wind turbine operations continuously. The available operational data for this case are presented in Table 2.

Table 2. Operational data.

| Operational Data | |
|---|---|
| Rotation speed | Min, Average, Max |
| Power output | Min, Average, Max |
| Energy output | Total, Diff |
| Nacelle position | Average |

**Status Data**: Last type of data collected by SCADA system provide information on the status of the system. Fault occurrences, maintenance actions and general information about turbine are included in the status data.

### 2.2 Generated Data

Besides the directly collected data, many additional features were generated with the aim of supplying inputs that carry information on incipient faults to improve the success of the fault detection system. The generated data consist of difference data, time series data, statistical data and knowledge-based data.

**Difference data:** The differences between the temperature measurements, operational data and wind speed measurements which may provide information on fault formations were calculated. For example, the differences between minimum, mean and maximum rotor powers, wind speeds, produced power values and the difference between temperature of generator rotor and generator stator are amongst these generated data. The total number of features generated by this way is 17.

**Time series data:** From 10 min to 60 min time delayed values of original features were generated to analyze time series characteristics. The number of time series features is 132.

**Statistical data:** Moving mean, standard variation and median values of original features up to 60 min were generated. 198 additional features were generated by this way.

**Knowledge-based data:** Based on the knowledge on wind turbine conversion systems, additional features were generated. The available power values were calculated using minimum, mean and maximum wind speed values for each interval. Also, the ratio of these values to the minimum, maximum and mean generated power were calculated. Sinusoidal components of nacelle position are also amongst the knowledge-based data.

By gathering the constructed features with the original features, a data set with 377 features was obtained. 22 of them are original features and the remaining 355 are generated features. The feature selection methodology is explained in the following part.

### 3. Feature Selection

### 3.1 Filter based feature selection

Filter approaches aim to rank features from the data alone [26]. Figure 1 presents the scheme of feature selection by filter methods. Using the training data, they determine which features are more relevant to the output of the.

Four filter-based feature selection methods were attempted in the initial simulations. They are Fisher Score [27], [28], Relief algorithm [29], Mutual information [30] and correlation-based feature selection [31]. The initial simulations showed that Fisher score and Relief methods had supplied effective results. Therefore, the detailed subset selection had been conducted using the results based on these two techniques.

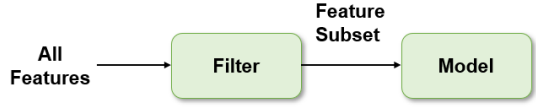

**Figure 1.** Filter based feature selection.

### 3.1.1 Fisher Score

Fisher method computes a score for each feature by calculating the ratio of the distance between data points in different classes to the distance between data points in same classes. The Fisher score of the $i^{th}$ feature is calculated as given in Equations 1-2;

$$F(i) = \frac{\sum_{k=1}^{c} n_k (\mu_k^i - \mu^i)^2}{(\sigma^i)^2} \quad (1)$$

$$(\sigma^i)^2 = \sum_{k=1}^{c} n_k (\sigma_k^i)^2 \quad (2)$$

Where, $n_k$ is the size of the $k-th$ class, $\mu_k^i$ and $\sigma_k^i$ are the mean and standard deviation of the $i-th$ feature when considering the samples of the $k-th$ class. $\mu_i$ and $\sigma^i$ are the mean and standard deviation of the whole data set corresponding to the $i-th$ feature. A higher Fisher score means that the informative value of the corresponding feature is also higher.

### 3.1.1 Relief Algorithm

Relief method [29] uses an Euclidian distance metric and nearest neighbor technique to rank the features based on their discriminative capabilities. It randomly selects instances from the training set and calculates a score based on the Equation 3.

$$R(i) = \frac{1}{2} \sum_{t=1}^{M} (\|x_{t,i} - NM(x_t)_i\| - \|x_{t,i} - NH(x_t)_i\|) \quad (3)$$

Where, $x_{t,i}$ is the value of feature i in the instance t. M is the number of the instances randomly selected from the data. $NH(x)$ are the nearest sample from the same class *('nearest hit')* and $NM(x)$ are the nearest sample from the opposite class *('nearest miss')* and $\|\cdot\|$ is the distance measurement. The algorithm calculates the discriminative success of each feature with respect to whether the feature differentiates two instances from the same class which is an undesired property and whether it differentiates two instances from opposite class which is a desired property [13].

### 3.2 Wrapper Method and Sequential Backward Floating Search

After selecting the most relevant features by the filter methods, a wrapper-based evaluation was employed to benefit from the advantages of both methods. Figure 2 shows wrapper feature selection principle. Various search algorithms can be used to decide the subsets to be used in wrapper models. Common search algorithms can be classified as exponential, sequential and randomized algorithms [32]. In exponential algorithms, number of subsets increases exponentially with the number of elements in the feature space. For example, exhaustive search is a kind of exponential search algorithms where all possible subsets of the feature space are used in the wrapper models to find the best combination, however even after the reduction of feature number by filter methods it is still computationally expensive in this case. Sequential search algorithms add or remove features sequentially. Sequential Forward Search, Sequential Backward Search, Sequential Floating Search Algorithms are amongst the main sequential search methods [33]. Randomized algorithms try to find optimum subsets by the use of randomness in their approach.

As a set of relevant features were obtained in the first part of this research, the remaining number of features became relatively smaller so the use of sequential search is appropriate to remove redundancies, evaluate mutual relations and increase the accuracy of the system. Sequential Backward Floating Search (SBFS) was selected for this aim. SBFS is a top down search procedure where the initial set starts by the whole feature set. The least significant feature is excluded in each step which is followed by conditional inclusions [33]. The search continues as long as the resulting subsets are better than the previously evaluated ones at that level.

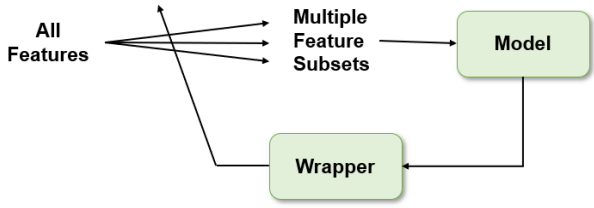

**Figure 2.** Wrapper based feature selection.

ANN models were implemented both as the induction algorithm of the wrapper model and as the final decision model. Mathematical foundations of ANNs can be found in [34]. The types of ANNs used are multilayer Multi-Layer Feed-Forward Neural Networks (MFFNN) where information flows from inputs to outputs without feedback connections and errors propagate from outputs to inputs. The ANN architectures consist of one input, one hidden and one output layer. The number of neurons in the hidden layer were changed from 2 to 15 to find architectures with high performance. Also, various activation functions were tried in the scope of this work. To prevent the algorithm from local minimums, multiple trials with random initial weights were held for each architecture.

### 3.4 Performance Metrics

The main performance metrics for classification problems can be listed as accuracy, specificity, recall, precision and f-score. Mathematical expressions of these metrics are presented in Equations 4-8.

$$Accuracy = (tp + tn)/(tp + tn + fp + fn) \quad (4)$$

$$Specificity = tn/(fp + tn) \quad (5)$$

$$Recall = tp/(tp + fn) \quad (6)$$

$$Precision = tp/(tp + fp) \quad (7)$$

$$FScore = \frac{2tp}{2tp + fp + fn} = 2 * \frac{Recall * Precision}{Recall + Precision} \quad (8)$$

Where, $tp$ is the true positives; number of correctly classified fault instances, $tn$ is the true negatives; number of correctly classified normal instances, $fp$ is the false positives; normal instances incorrectly predicted as fault instances and $fn$ is the false negatives; fault instances incorrectly classified as normal instances.

The case observed in this paper is a two-class classification problem where the output can either belong to the normal or the faulty class. Due to the natural characteristics of fault detection systems, there is a significant imbalance ratio which means number of the elements in the majority class is much higher than the elements in the minority class. In our data set, there are 223 instances where the status of the turbine shows a generator heating fault, whereas there are more than 50000 normal operation instances without generator heating faults. In such cases, accuracy and specificity might not be appropriate performance metrics as the nature of the application needs a high rate of correctly classified minority class samples [35]. However, even if there are no correctly classified fault instances, the accuracy and the specificity would still be high due to the high number of correctly classified normal instances. Because of this reason, recall, precision and f-score which is the harmonic mean of the former two were used as they supply information on the success of the models.

### 4. Results and Discussions

To evaluate the feature construction and hybrid feature selection method proposed in this paper, a comparative analysis was conducted. Generator heating faults were attempted to be detected in two different ways. Firstly, the method presented in this paper was used. For the second case, a heuristic feature construction and selection method mainly by using the original features and some additional features constructed by expert knowledge was used. The constructed features in the heuristic case include the knowledge-based features, differences, moving variance and moving mean values belong to the generator temperature sensors. In order to improve the performance of the model as much as possible, a backward floating search amongst these features was also applied in the heuristic case. All the analyses were performed in MATLAB environment.

Table 3 presents the most informative features determined by the proposed and heuristic methods. As can be seen from the table, when the proposed approach is used, only 2 of the resulting features are amongst the original features which are the maximum power output and the minimum wind speed. Whereas rest of them are the constructed features.

**Table 3.** Selected features by the proposed and heuristic methods.

| Heuristic Method |
|---|
| Minimum wind speed |
| Mean wind speed |
| Minimum rotor speed* |
| Mean rotor speed |
| Minimum power output |
| Maximum power output* |
| Generator rotor temperature |
| Generator stator temperature |
| 60-min moving variance of gen. stator temperature |
| 60-min moving variance of gen. rotor temperature |
| **Proposed Method** |
| Gen. stator temperature – Gen. rotor temperature |
| Gen. rotor temperature – Transformer temperature |
| Gen. rotor temperature – Nacelle temperature |
| 60-min median of max power output |
| 30-min median of max power output |
| Max power output* |
| 30-min median of max power output |
| 60-min mean of max power output |
| Minimum rotor speed* |
| Max available power from wind |

*Mutual features selected by two methods.

The selected features and the results belong to two different cases show the proposed method's effective handling of mutual relations between features. The most important feature that was found by the proposed method is the difference between the generator stator temperature and the generator rotor temperature. This feature was also tried in the heuristic approach. However, in that case instead of increasing the classification performance, it has even reduced it. This situation shows that this feature is very informative only when it is used along with the other features obtained by the hybrid approach.

By using the resulting features obtained in two methods, ANNs were trained. 70% of the data were used in the training phase and rest of the data were used in the test phase. Various ANNs were trained as by changing the activation functions, number of hidden neurons and initial weights. The results show that the proposed approach effectively improves the performance of the fault detection algorithm. Figure 3 shows the values of recall, precision and f1 performance metrics for the most successful networks in both approaches. Best balanced scores were taken into consideration. Accuracy and specificity scores are greater than 0.98 in both cases, however they are not discriminative metrics in this problem due to the high imbalance ratio of fault and no-fault classes.

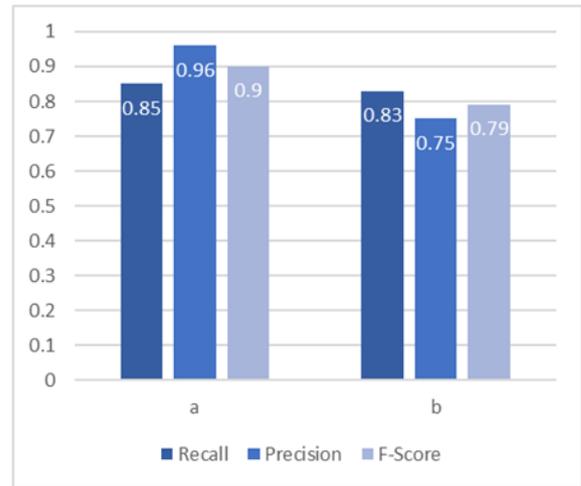

**Figure 3.** Performance metrics for a) proposed b) heuristic methods.

As it is seen in Figure 3, the proposed method is effective on improving the success of the detection. There is a slight increase in the recall score which is 0.83 in the heuristic case and 0.85 in the proposed method. A significant improvement was observed in the precision score which increased from 0.75 to 0.96. This result mostly arose from the decrease in the number of false positives which express the number of normal instances incorrectly classified as fault instances. In this problem, the decrease in the false positives means there are less false fault alarms when the classification is performed by the proposed method. In the approximately 3 months of test period, false fault alarm duration obtained in the heuristic method was 210 min whereas it was 30 min in the proposed method. As a result, this work contributes to the solution of one of the main challenges of wind turbine fault detection using SCADA data.

## 5. Conclusion

An intelligent method for improving the fault detection performance in wind turbines is proposed in this paper. The algorithm is comprised of two main parts, a feature construction step and a feature selection step. In the feature construction part, in addition to the original features directly collected from SCADA system, various features were generated with the aim of providing stronger fault indications. In the feature selection part, filter and wrapper methods were implemented in a hybrid manner. To better analyze the results obtained by the proposed method, a heuristic feature selection was also implemented to the same problem.

A major advantage of this work is that it helps better use of SCADA data in wind turbines. As these kinds of non-fatal errors show less indications, it is a challenging task to detect them only from the SCADA system which was not initially developed for fault detection purposes. Moreover, further difficulties were present in this work due to the absence of some types of data which are commonly obtainable in many SCADA systems. One of the biggest problems from the fault detection attempts from SCADA data is high number of false alarm rates. The methods proposed in this paper reduced false alarm rates significantly from 210 to 30 minutes. They were effective both in increasing the recall and precision metrics which are the informative factors for highly imbalanced datasets. For future works, generalization of these results to a larger dataset can be focused on.

This cost-effective approach would be instrumental in the design of AI based condition monitoring systems for wind energy industry which is gaining increasing importance due to the increasing global demands from the sustainable energy systems.

## References

[1] W. Qiao and D. Lu, "A Survey on Wind Turbine Condition Monitoring and Fault Diagnosis - Part II: Signals and Signal Processing Methods," *IEEE Trans. Ind. Electron.*, vol. 62, no. 10, pp. 6546–6557, 2015.

[2] A. Abouhnik and A. Albarbar, "Wind turbine blades condition assessment based on vibration measurements and the level of an empirically decomposed feature," *Energy Convers. Manag.*, vol. 64, pp. 606–613, 2012.

[3] Y. Qu, E. Bechhoefer, D. He, and J. Zhu, "A New Acoustic Emission Sensor Based Gear Fault Detection Approach," *Int. J. Progn. Heal. Manag.*, vol. 4, pp. 1–14, 2013.

[4] O. M. Bouzid, G. Y. Tian, K. Cumanan, and D. Moore, "Structural Health Monitoring of Wind Turbine Blades : Acoustic Source Localization Using Wireless Sensor Networks," *J. Sensors*, vol. 2015, p. 11, 2014.

[5] M. Moradi and S. Sivoththaman, "MEMS multisensor intelligent damage detection for wind turbines," *IEEE Sens. J.*, vol. 15, no. 3, pp. 1437–1444, 2015.

[6] A. Coscetta *et al.*, "Wind Turbine Blade Monitoring with Brillouin-Based Fiber-Optic Sensors," *J. Sensors*, vol. 2017, 2017.

[7] M. Schlechtingen, I. F. Santos, and S. Achiche, "Wind turbine condition monitoring based on SCADA data using normal behavior models. Part 1: System description," *Appl. Soft Comput.*, vol. 13, no. 1, pp. 259–270, 2013.

[8] W. Yang, R. Court, and J. Jiang, "Wind turbine condition monitoring by the approach of SCADA data analysis," *Renew. Energy*, vol. 53, pp. 365–376, 2013.

[9] K. Leahy, R. L. Hu, I. C. Konstantakopoulos, C. J. Spanos, and A. M. Agogino, "Diagnosing wind turbine faults using machine learning techniques applied to operational data," *2016 IEEE Int. Conf. Progn. Heal. Manag.*, pp. 1–8, 2016.

[10] Y. Zhao, D. Li, A. Dong, J. Lin, D. Kang, and L. Shang, "Fault prognosis of wind turbine generator using SCADA data," *NAPS 2016 - 48th North Am. Power Symp. Proc.*, pp. 0–5, 2016.

[11] P. B. Dao, W. J. Staszewski, T. Barszcz, and T. Uhl, "Condition monitoring and fault detection in wind turbines based on cointegration analysis of SCADA data," *Renew. Energy*, pp. 1–16, 2017.

[12] F. Castellani, D. Astolfi, P. Sdringola, S. Proietti, and L. Terzi, "Analyzing wind turbine directional behavior: SCADA data mining techniques for efficiency and power assessment," *Appl. Energy*, vol. 185, pp. 1076–1086, 2017.

[13] H. Liu, "Computational Methods of Feature Selection," Chapman&Hall, 2007.

[14] Y. Li and X. Zeng, "Sequential multi-criteria feature selection algorithm based on agent genetic algorithm," *Appl. Intell.*, vol. 33, no. 2, pp. 117–131, 2010.

[15] G. Roffo, S. Melzi, and M. Cristani, "Infinite feature selection," *Proc. IEEE Int. Conf. Comput. Vis.*, vol. 2015 Inter, pp. 4202–4210, 2015.

[16] G. Roffo and S. Melzi, "Online Feature Selection for Visual Tracking," *Bmvc*, 2016.

[17] Y. Wang, L. Feng, and Y. Li, "Two-step based feature selection method for filtering redundant information," *J. Intell. Fuzzy Syst.*, vol. 33, no. 4, pp. 2059–2073, 2017.

[18] Y. Xue, L. Zhang, B. Wang, Z. Zhang, and F. Li, "Nonlinear feature selection using Gaussian kernel SVM-RFE for fault diagnosis," *Appl. Intell.*, pp. 1–26, 2018.

[19] X. Huang, L. Zhang, B. Wang, F. Li, and Z. Zhang, "Feature clustering based support vector machine recursive feature elimination for gene selection," *Appl. Intell.*, vol. 48, no. 3, pp. 594–607, 2018.

[20] J. Tang, S. Alelyani, and H. Liu, "Feature Selection for Classification: A Review," *Data Classif. Algorithms Appl.*, pp. 37–64, 2014.

[21] R. Kohavi and G. H. John, "Wrappers for feature subset selection," *Artif. Intell.*, vol. 97, no. 1–2, pp. 273–324, 1997.

[22] M. Dash and H. Liu, "Feature Selection for Classification," *IDA ELSEVlER Intell. Data Anal.*, vol. 1, no. 97, pp. 131–156, 1997.

[23] A. Kusiak and W. Li, "The prediction and diagnosis of wind turbine faults," *Renew. Energy*, vol. 36, no. 1, pp. 16–23, 2011.

[24] R. L. Hu, K. Leahy, I. C. Konstantakopoulos, D. M. Auslander, C. J. Spanos, and A. M. Agogino, "Using domain knowledge features for wind turbine diagnostics," *Proc. - 2016 15th IEEE Int. Conf. Mach. Learn. Appl. ICMLA 2016*, pp. 300–305, 2017.

[25] K. Leahy, R. L. Hu, I. C. Konstantakopoulos, C. J. Spanos, A. M. Agogino, and D. T. J. O'Sullivan, "Diagnosing and Predicting Wind Turbine Faults from SCADA Data Using Support Vector Machines," *Int. J. Progn. Heal. Manag.*, vol. 9, no. 1, pp. 1–11, 2018.

[26] H. Liu and H. Motoda, Feature Extraction, Construction and Selection: A Data Mining Perspective. 1998.

[27] R. O. Duda, P. E. Hart, and D. G. Stork, *Pattern Classification*. John Wiley & Sons, New York, 2001.

[28] Q. Gu, Z. Li, and J. Han, "Generalized Fisher Score for Feature Selection," 2012.

[29] M. Robnik-Sikonja and I. Kononenko, "Theoretical and empirical analysis of ReliefF and RRiefF," *J. Mach. Learn. Res.*, vol. 53, pp. 23–69, 2003.

[30] H. Peng, F. Long, and C. Ding, "Feature selection based on mutual information: Criteria of Max-Dependency, Max-Relevance, and Min-Redundancy," *IEEE Trans. Pattern Anal. Mach. Intell.*, vol. 27, no. 8, pp. 1226–1238, 2005.

[31] M. Hall, "Correlation-based Feature Selection for Machine Learning," 1999.

[32] L. C. Molina, L. Belanche, À. Nebot, J. Girona, and C. N. C, "Feature Selection Algorithms A Survey and Experimental Evaluation" *Data Mining, 2002. ICDM 2002. Proceedings. 2002 IEEE Int. Conf.*, pp. 306–313, 2002.

[33] P. Pudil, J. Novovičová, and J. Kittler, "Floating search methods in feature selection," *Pattern Recognit. Lett.*, vol. 15, no. 11, pp. 1119–1125, 1994.

[34] S. Haykin, *Neural Networks and Learning Machines*. 2009.

[35] N. V. Chawla, "Data Mining for Imbalanced Datasets: An Overview," *Data Min. Knowl. Discov. Handb.*, pp. 875–886, 2009.